\documentclass[a4paper,fleqn]{cas-sc}

\usepackage[square,numbers]{natbib}
\usepackage{hyperref}
\usepackage{subcaption}  
\def\tsc#1{\csdef{#1}{\textsc{\lowercase{#1}}\xspace}}
\tsc{WGM}
\tsc{QE}

\begin{document}
\let\WriteBookmarks\relax
\def\floatpagepagefraction{1}
\def\textpagefraction{.001}

\shorttitle{}    

\shortauthors{}  

\title [mode = title]{Explainable AI for Sentiment Analysis of Human Metapneumovirus (HMPV) Using XLNet}  


%

\author[1]{Md. Shahriar Hossain Apu}
\ead{1901036@iot.bdu.ac.bd}

\author[2]{Md Saiful Islam} 
\ead{2002047@icte.bdu.ac.bd}

\author[3]{Tanjim Taharat Aurpa}\cormark[1]
\ead{aurpa0001@bdu.ac.bd}

\address[1]{Department of IoT and Robotics Engineering, Bangabandhu Sheikh Mujibur Rahman Digital University, Bangladesh}

\address[2]{Department of Educational Technology and Engineering, Bangabandhu Sheikh Mujibur Rahman Digital University, Bangladesh}

\address[3]{Department of Data Science and Engineering, Bangabandhu Sheikh Mujibur Rahman Digital University, Bangladesh}

\cortext[1]{Coresponding Author: Tanjim Tahrat Aurpa}


\begin{abstract}
In 2024, the outbreak of Human Metapneumovirus (HMPV) in China, which later spread to the UK and other countries, raised significant public concern due to its potential impact on vulnerable populations. While HMPV typically causes mild symptoms, its effects on the elderly and immunocompromised individuals prompted health authorities to emphasize preventive measures. Moreover, continuous monitoring of respiratory viruses like HMPV remains important, as new factors (such as emerging variants) could alter their behavior over time. These factors have led to mixed public reactions, with some individuals expressing anxiety while others exhibit carelessness regarding the virus. This paper explores how sentiment analysis can enhance our understanding of public reactions to HMPV by analyzing data from social media platforms like YouTube. It highlights the importance of tracking public sentiment—ranging from fear to trust—to guide health messaging, inform policies, address misinformation, and encourage compliance with preventive measures during outbreaks.  This study focuses on the use of sentiment analysis to understand public reactions to HMPV during the 2024 outbreak. The research applies advanced transformer models, particularly XLNet, achieving an accuracy of 93.50\% in sentiment classification tasks. Additionally, We incorporate explainable AI (XAI) through SHAP to provide transparency in how the model identifies key factors influencing public sentiment.

\end{abstract}



\maketitle

\section{Introduction}

Human Metapneumovirus (HMPV) is a respiratory virus that causes cold-like symptoms such as cough, sore throat, runny nose, and wheezing. While most infections remain mild, severe cases can lead to pneumonia or exacerbate preexisting chronic conditions, including asthma and chronic obstructive pulmonary disease (COPD). High-risk groups, such as young children under five, elderly individuals over 65, and immunocompromised patients, are particularly susceptible to severe illness. HMPV is primarily transmitted through respiratory droplets and contaminated surfaces, with most cases resolving spontaneously. However, critical cases may require medical intervention \cite{clevelandclinic_hmpv}. 

HMPV infections exhibit a seasonal pattern, with peak cases occurring during late winter and early spring. This trend is attributed to the virus's enhanced survival in colder conditions and increased human contact in indoor settings \cite{bbc_hmpv}. Recently, a significant rise in HMPV cases has been observed across the Northern Hemisphere, particularly in countries like China, the United States, and the United Kingdom, since late 2024. Despite this surge, public health systems have remained stable, with the World Health Organization (WHO) actively monitoring the global situation \cite{who_hmpv_2025}.  

Given the emerging concern around HMPV and its increasing global prevalence, there is a lack of comprehensive research focused on public sentiment regarding the virus. Traditional data sources for sentiment analysis, such as surveys or medical reports, are limited, especially in capturing real-time public discourse. To bridge this gap, we turn to YouTube, a popular platform where users often discuss personal experiences and concerns. YouTube comments provide a unique, real-time window into public sentiment and are rich with diverse, spontaneous reactions to health-related topics. As this is the first research of its kind on HMPV-related sentiment analysis, we chose to collect comments specifically related to HMPV in order to explore and understand the public’s feelings and reactions.

As respiratory illnesses become more prevalent during winter, understanding public sentiment surrounding HMPV is crucial. Sentiment analysis of public discourse can help health organizations identify widespread concerns, address misinformation, and improve communication strategies. Sentiment analysis, a key area of natural language processing (NLP) \cite{JIM2024100059}, involves classifying text into sentiment categories such as positive, negative, or neutral. Traditional sentiment analysis methods, including rule-based and statistical models, often struggle with capturing contextual and nuanced meanings in text \cite{MAO2024102048}.  

Recent advances in transformer-based models, such as BERT, XLNet, and RoBERTa, have significantly improved NLP tasks by providing superior contextual understanding and performance. These models are particularly effective in sentiment analysis, offering enhanced accuracy in interpreting complex textual data. To improve transparency and trust in sentiment classification, Explainable AI (XAI) techniques, such as SHAP (Shapley Additive Explanations), can be integrated to interpret and visualize model decisions.  

XLNet, an advanced autoregressive transformer model, has demonstrated strong performance in various NLP tasks, including Named Entity Recognition \cite{yan2021named}, Sentiment Analysis \cite{sweidan2021sentence}, and Emotion Detection \cite{shen2021dialogxl}. It is an enhanced version of Transformer-XL designed to overcome the text sequence length limitations of conventional transformers. By combining autoregressive and autoencoding language models, XLNet effectively captures bidirectional context while minimizing their respective shortcomings. Its architecture employs a unique attention mechanism known as Two-Stream Self-Attention, along with positional encoding. Unlike BERT, which follows a masked language model approach, XLNet utilizes Permutation Language Modeling (PLM) during pretraining, allowing it to consider all possible token sequences. This approach has enabled XLNet to outperform BERT in 20 distinct NLP benchmarks, demonstrating its superior performance in downstream tasks.

This research aims to develop a sentiment analysis framework for HMPV-related discourse using transformer-based architectures and XAI techniques. By leveraging a preprocessed and annotated dataset of user comments, the system will classify sentiment with high accuracy while providing interpretable explanations for its predictions.

The main contributions of this study are as follows:  
\begin{itemize}
    \item Development of a robust sentiment analysis pipeline for HMPV-related text using state-of-the-art NLP techniques.
    \item Integration of SHAP to enhance the transparency and interpretability of sentiment classification models.
    \item Insights into public sentiment on HMPV, aiding in improved health communication and policy development.
\end{itemize}

By integrating cutting-edge machine learning models with explainability, this study bridges the gap between computational advancements and their real-world applications in public health. The findings of this research can facilitate more effective and informed decision-making in response to HMPV outbreaks.

The remainder of this paper is organized as follows: Section 2 provides a detailed review of the literature, exploring relevant works in sentiment analysis, HMPV-related research, and the application of transformer-based models. Section 3 outlines the methodology employed in this study, including the data collection process, model architecture, and integration of Explainable AI (XAI) techniques. Section 4 presents the results, showcasing the performance of the sentiment analysis framework and its interpretability. In Section 5, the discussion is centered around the findings, highlighting the implications of public sentiment and the contributions of the study. Finally, Section 6 concludes the paper with a summary of key findings and outlines potential directions for future research.

\section{Literature Review}

Human Metapneumovirus (hMPV) was discovered in 2001 and is a leading cause of respiratory infections, particularly among young children, the elderly, and immune-compromised individuals\cite{PANDA201445}. This virus is responsible for 5-10\% of pediatric hospitalizations due to acute respiratory infections and can lead to severe conditions like bronchiolitis and pneumonia. The symptoms of hMPV infection often resemble those of respiratory syncytial virus, and RT-PCR is the preferred diagnostic method due to the virus's slow growth in cell culture. Although vaccine candidates are under development, no vaccines are currently commercially available, and research continues to improve the understanding and treatment strategies for hMPV.

A case report by \cite{microorganisms13010073} discusses a 68-year-old immunocompetent male who developed severe pneumonia caused by hMPV. Despite the absence of significant comorbidities, the patient required hospitalization due to worsening respiratory symptoms. Diagnosis was confirmed with multiplex RT-PCR, and imaging revealed viral pneumonia. The patient fully recovered with supportive care, underscoring the importance of molecular diagnostics in accurate diagnosis, reducing unnecessary antibiotic use, and highlighting the potential for future vaccines such as IVX-A12.

Kannappan et al.\cite{article12} examine the growing significance of sentiment analysis in the digital landscape. Their study highlights its role in website development, social media profile creation, and managing digital platforms. Sentiment analysis aids in addressing customer inquiries, evaluating product feedback, and safeguarding a company's reputation by ensuring positive reviews. The research explores the synergy between Natural Language Processing (NLP) and Machine Learning (ML), focusing on how NLP tools process human language and how ML, particularly with Python, facilitates effective sentiment analysis.

Kavitha et al.\cite{9823708} explore the application of sentiment analysis on social media data using NLP and ML techniques. Their study emphasizes the importance of extracting insights from user-generated content like tweets and blogs, showing how ML algorithms such as Random Forest and Logistic Regression can effectively process and analyze sentiment. This research demonstrates the potential of combining NLP and ML to derive meaningful conclusions from social media data and enhance decision-making.

Srivastava et al.\cite{9242618} review various approaches to sentiment analysis within NLP, focusing on techniques that detect the emotional tone of text. With the exponential growth of online content, including text, photos, audio, and video, their study illustrates how sentiment analysis can extract valuable insights. The authors examine widely-used methods such as Naïve Bayes, Support Vector Machines (SVM), and the lexicon-based approach in NLP, and discuss challenges in sentiment analysis, predicting greater accessibility for smaller businesses and the public as technology evolves.

Jim et al.\cite{JIM2024100059} present a comprehensive review of recent advancements and challenges in sentiment analysis, a critical area within NLP. They examine its application in classifying textual data as positive, negative, or neutral, providing businesses with crucial insights into customer emotions. The study delves into various domains, pre-processing techniques, datasets, and evaluation metrics that contribute to sentiment analysis, as well as the roles of Machine Learning, Deep Learning, and Large Language Models. They propose future research directions to address the challenges and limitations identified in state-of-the-art studies.

Gunasekaran\cite{unknown1} reviews various sentiment analysis techniques within NLP, including lexicon-based, machine learning, deep learning, and hybrid approaches. This research highlights the importance of sentiment analysis in customer feedback analysis, marketing, and politics. Using Twitter as a case study, it discusses applications across different sectors and offers a comparative analysis of techniques, datasets, and metrics aimed at improving the accuracy and efficiency of sentiment analysis.

Chong et al.\cite{7351837} explore sentiment analysis on tweets using NLP techniques, focusing on three key steps: subjectivity classification, semantic association, and polarity classification. They employ sentiment lexicons and grammatical relationships, achieving superior results over traditional sentiment analysis tools. Their findings emphasize the importance of tailored NLP approaches for analyzing sentiment in social media contexts.

Jain et al.\cite{Jain2023} focus on real-time sentiment analysis using multimedia inputs such as audio, video, and text to interpret emotions. They compare techniques like Support Vector Machines (SVMs), Bayesian Networks, and Convolutional Neural Networks (CNNs) for sentiment classification. Their research leads to the development of a real-time sentiment analysis system to help users assess daily attitudes and receive relevant recommendations.

Wankhade et al.\cite{Wankhade2022} provide a comprehensive survey on sentiment analysis, reviewing methods, applications, and challenges. The study explores how sentiment analysis can gather and interpret opinions from internet-based platforms like social media and blogs. It compares various techniques and addresses challenges in accurately determining sentiment polarity, offering suggestions for future research to improve the effectiveness and accuracy of sentiment analysis.

The reviewed literature highlights the broad range of applications and techniques in sentiment analysis, particularly in the domains of customer feedback, social media analysis, and emotion detection from multimedia data. The ongoing advancements in NLP and ML offer significant potential for improving sentiment analysis methods, with future research expected to address the current challenges and further refine these technologies for diverse real-world applications.

\section{Methodology}
In this section, the preliminaries of our methodology and
the proposed approach used here is discussed. The major part
of the methodology of this research is explained below:
\subsection{Dataset}
The dataset we used in our study is a collection of comments gathered from different news channels on YouTube. After data scraping, we collected 15,300 comments. After applying various preprocessing steps, the dataset was reduced to 9,758 comments. While preparing the dataset, we also labeled the comments with sentiment labels and scores. 
The dataset is publicly available and can be accessed at the following link \cite{Hossain2025Dataset}.

\subsubsection{Data Preprocessing}
Data preprocessing is a critical phase for ensuring accurate evaluation and improving sentiment analysis outcomes \cite{Krouska2016Preprocessing}. During our preprocessing efforts, we conducted manual cross-checks to eliminate irrelevant content and reduce noise, enhancing the dataset's reliability and integrity.

Most raw data  contains noise, irrelevant information, and inconsistencies that may affect negatively impact machine learning model performance \cite{Duong2021Preprocessing}. As such, effective preprocessing is  very important.

\begin{figure}[h] \centering \includegraphics[width=0.8\textwidth]{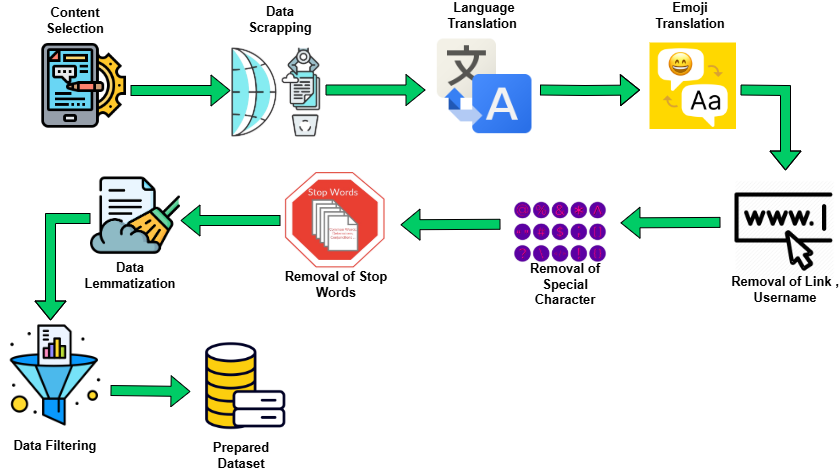} \caption{Data Collection and Preprocessing.} \label{fig:dataPreprocess} \end{figure}

From our YouTube comment scraping, we observed significant amounts of spam, advertisements, and irrelevant discussions that provided no value for our analysis. Additionally, the dataset also included with numerous grammatical errors, typing errors, special characters, and URLs. These elements, collectively considered as noise and needed cleaning before application of
machine learning models to avoid poor evaluations \cite{Bhuiyan2017YouTube}. To address these issues, the following preprocessing steps were implemented and preprocessing steps shown in figure: \ref{fig:dataPreprocess}:

\textbf{Language Translation:}
Many comments were written in many languages. To apply machine learning algorithms, we had to translate all those into English.

\textbf{Emoji Translation:}
Emojis tend to convey emotions that can be missed with plain text. To capture this emotion, emojis were translated into their textual equivalents to match the sentiment that they convey.

\textbf{Removal of Links, Emails, and Usernames:}
Any hyperlinks, email addresses, or usernames found in the comments were removed to enhance the dataset's focus and quality.

\textbf{Removal of Special Characters:}
Special characters in the dataset were eliminated to simplify and standardize the data.

\textbf{
Stopword Removal:}
Common stopwords that did not contribute meaningfully to sentiment analysis were removed to reduce noise and improve model performance.

\textbf{Lemmatization:}Lemmatization was performed to convert the words into their base or root forms, maintaining the consistency and reducing dimensionality in the dataset. The task of lemmatization is to determine the basic form of a given word, which enhances the quality of text analysis\cite{Skorkovska2012Lemmatization}.

By applying these preprocessing techniques, we ensured the dataset was clean, relevant and ready for analysis, significantly enhancing the overall performance of our machine learning models.

\subsection{Data Labeling}
\subsubsection{Sentiment Label Identification} For Sentiments Identification we used VADER,Which is lexion and rule based identification method \cite{Hutto2014VADER}, VADER is well suited for analyzing media texts and informal language due to its ability to capture sentiment intensity accurately.
This tool classify each text such as Positive,Negative and Neutral and also provide sentiment scores which helps us for our evaluations.

\subsection{Model Building, Training and Evaluation } 
After completing Data Collection and Data Preprocessing, We build our model and trained it. Here is an overview of our proposed model.

\subsubsection{XLNet Model}
XLNet is an advanced natural language processing model \cite{yang2020xlnetgeneralizedautoregressivepretraining}developed as an extension of the Transformer-XL architecture, introduced to overcome the limitations of previous models, such as BERT, by handling longer text sequences and leveraging permutation-based training. It combines the strengths of both autoencoding and autoregressive models, capturing bidirectional context while also maintaining the autoregressive nature of language modeling. This hybrid approach allows XLNet to outperform BERT in various NLP tasks, including sentiment analysis, text classification, and question answering, by improving context understanding and sequence generation\cite{article1}.

XLNet’s architecture incorporates bidirectional context and positional encoding while utilizing a novel attention mechanism known as Two-Stream Self-Attention. Like BERT, XLNet also follows a two-step process: pretraining and fine-tuning, but XLNet introduces a key innovation—Permutation Language Modeling (PLM)—which considers all possible permutations of the input sequence during pretraining. This unique approach enables XLNet to capture more contextual information and overcome BERT’s limitation of only considering left-to-right or right-to-left context.

\begin{figure}[h]
    \centering
    \includegraphics[width=0.8\linewidth]{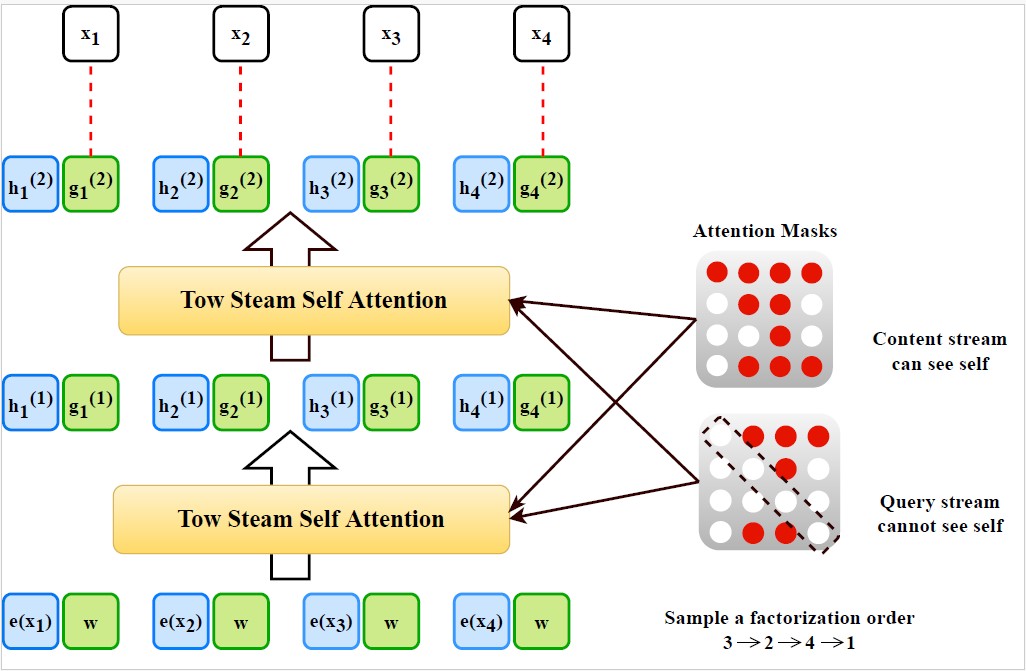}
    \caption{Proposed XLNet Model}
    \label{xlnetmodell}
\end{figure}

\textbf{Pre-Training XLNet:}
In the pre-training phase, XLNet is trained on large corpora of unlabeled text using Permutation Language Modeling (PLM), which differs from BERT's Masked Language Modeling. PLM randomly permutes the sequence of input tokens, and XLNet learns to predict the next token in the sequence based on these permutations. Figure-\ref{xlnetmodell} shows the working strategy of the permutation language model(PLM).
This approach allows XLNet to utilize all possible context within the sequence, offering a more flexible and powerful understanding of the text. Unlike BERT, which relies on a fixed left-to-right or right-to-left direction, XLNet’s permutation-based model enhances its performance on downstream tasks by better capturing the relationship between tokens.

\textbf{Fine-Tuning XLNet:}
After pretraining, XLNet is fine-tuned on specific labeled datasets for downstream tasks, including text classification, question answering, and sentiment analysis. This phase adapts the general language understanding achieved in pretraining to perform tasks that require more targeted knowledge and interpretation of text.

\subsection{Hyperparameter Tuning}

Hyperparameter tuning involves adjusting the model’s hyperparameters, which significantly impact its performance, including its capacity, convergence speed, and ability to capture data patterns effectively. Experimenting with different combinations of hyperparameters and evaluating the results are essential steps in optimizing the model for a specific task.

For XLNet, hyperparameter tuning affects model performance by influencing its learning dynamics and sensitivity to input data. The model’s hyperparameters, such as learning rate, batch size, number of epochs, and sequence length, are fine-tuned to achieve the best possible performance. The following table shows the optimal hyperparameters found through experimentation.

\begin{table}[h!]
\centering
\caption{This table lists the model parameters and their corresponding values for XLNet.}
\label{tab:hyperparameters}
\begin{tabular}{|l|c|}
\hline
\textbf{Hyperparameters} & \textbf{XLNet Values} \\ \hline
Learning rate (AdamW)    & 2e-04                \\ \hline
Max length               & 50                   \\ \hline
Batch size               & 12                   \\ \hline
Verbose                  & 1                    \\ \hline
Epoch                    & 10                   \\ \hline
\end{tabular}
\end{table}

\subsection{Model Evaluation}

For the evaluation of our model, this research utilized a variety of evaluation metrics, such as accuracy and micro and macro average F1 scores. The calculation of the confusion matrix played a pivotal role in determining these metrics. The following equations were used to compute the accuracy and micro and macro average F1 scores:

\begin{equation}
\text{Accuracy} = \frac{\text{TP} + \text{TN}}{\text{TP} + \text{TN} + \text{FP} + \text{FN}}
\end{equation}

\begin{equation}
\text{Micro F1 Score} = \frac{\sum \text{TP}}{\sum \text{TP} + \frac{1}{2} \left(\sum \text{FN} + \sum \text{FP}\right)}
\end{equation}

\begin{equation}
\text{Macro F1 Score} = \frac{\sum_{i=1}^{\text{No of classes}} \text{F1 Score}_i}{\text{No of classes}}
\end{equation}

Here, TP, TN, FP, and FN indicate True Positive, True Negative, False Positive, and False Negative, respectively.

Additional equations were also used in this research to calculate Precision, Recall, Specificity, and Error Rate, as given below:

\begin{equation}
\text{Precision} = \frac{\text{TP}}{\text{TP} + \text{FP}} \times 100\%
\end{equation}

\begin{equation}
\text{Sensitivity/Recall} = \frac{\text{TP}}{\text{TP} + \text{FN}} \times 100\%
\end{equation}

\begin{equation}
\text{Specificity} = \frac{\text{TN}}{\text{TN} + \text{FP}}
\end{equation}

\begin{equation}
\text{Error Rate} = \frac{\text{FP} + \text{FN}}{\text{TP} + \text{TN} + \text{FP} + \text{FN}}
\end{equation}

\subsection{Model Explainability}

SHAP (SHapley Additive exPlanations) is a unique and valuable tool in our field, as it has been utilized to explain and understand our model’s important features for sentiment analysis of comments regarding the HMPV virus. This popular framework for interpreting the output of machine learning models assigns each feature an importance value for a given prediction based on Shapley values from cooperative game theory. These values provide a fair distribution of payoff among players (or features) in a coalition. 

In the context of sentiment analysis, SHAP’s contribution lies in its ability to elucidate how different features of the comments, such as specific words or phrases, contribute to the model’s sentiment predictions. By leveraging SHAP values, we can gain insights into the model’s decision-making process, particularly:

\begin{itemize}
    \item \textbf{Feature Importance:} SHAP highlights which features (e.g., words, phrases, or patterns in the comment) are most influential in determining the sentiment, whether positive, negative, or neutral.
    \item \textbf{Model Transparency:} The utilization of SHAP in this work enhances the transparency of our sentiment analysis model. By allowing us to understand why certain sentiment predictions are made, SHAP builds trust in the model’s reliability, making it a valuable tool for practitioners in the field.
    \item \textbf{Error Analysis:} SHAP aids in diagnosing errors with pinpoint precision, identifying features in the comments that lead to incorrect sentiment predictions and facilitating targeted model improvements.
\end{itemize}

\begin{figure}[h]
    \centering
\includegraphics[width=.7\linewidth]{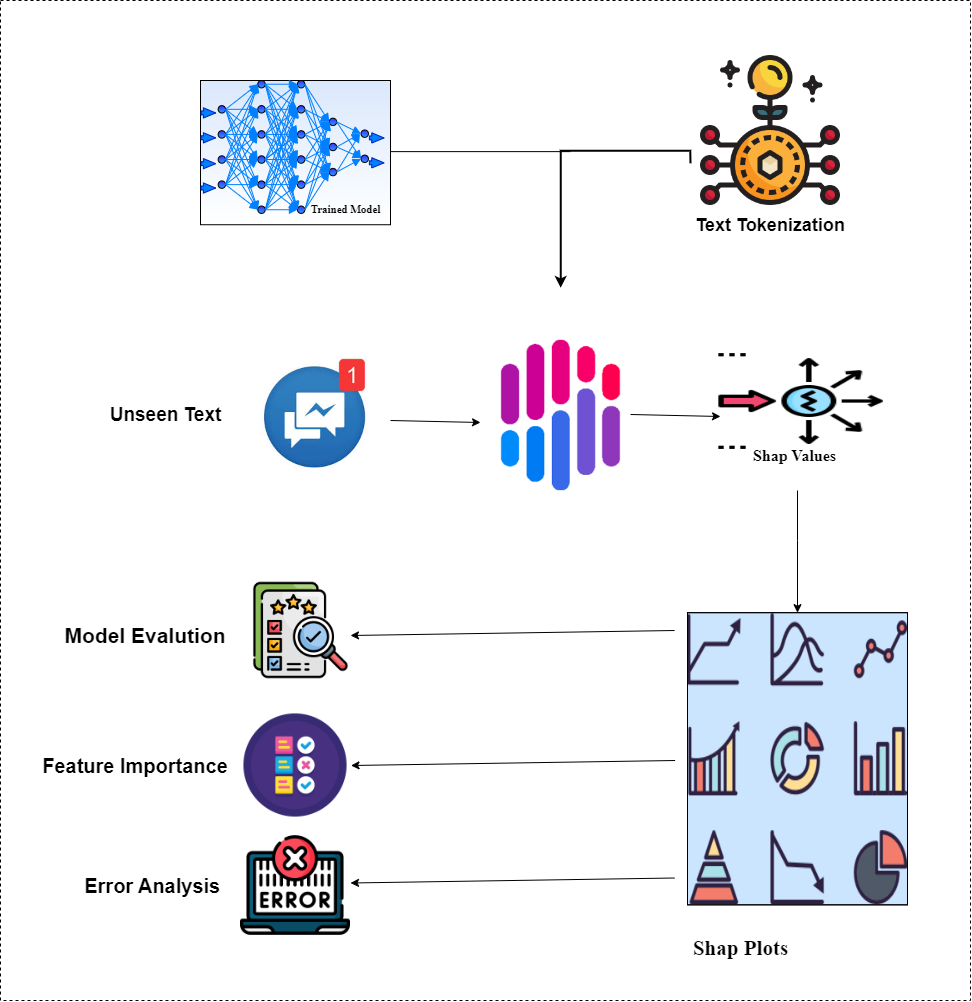}
    \caption{The working process of SHAP}
    \label{SHAP}
\end{figure}

Figure \ref{SHAP} illustrates the working principle of the SHAP algorithm, showing how individual features contribute to sentiment classification for comments on the HMPV virus.

\section{Proposed Approach}
\begin{figure}[h]
    \centering
    \includegraphics[width=0.7\linewidth]{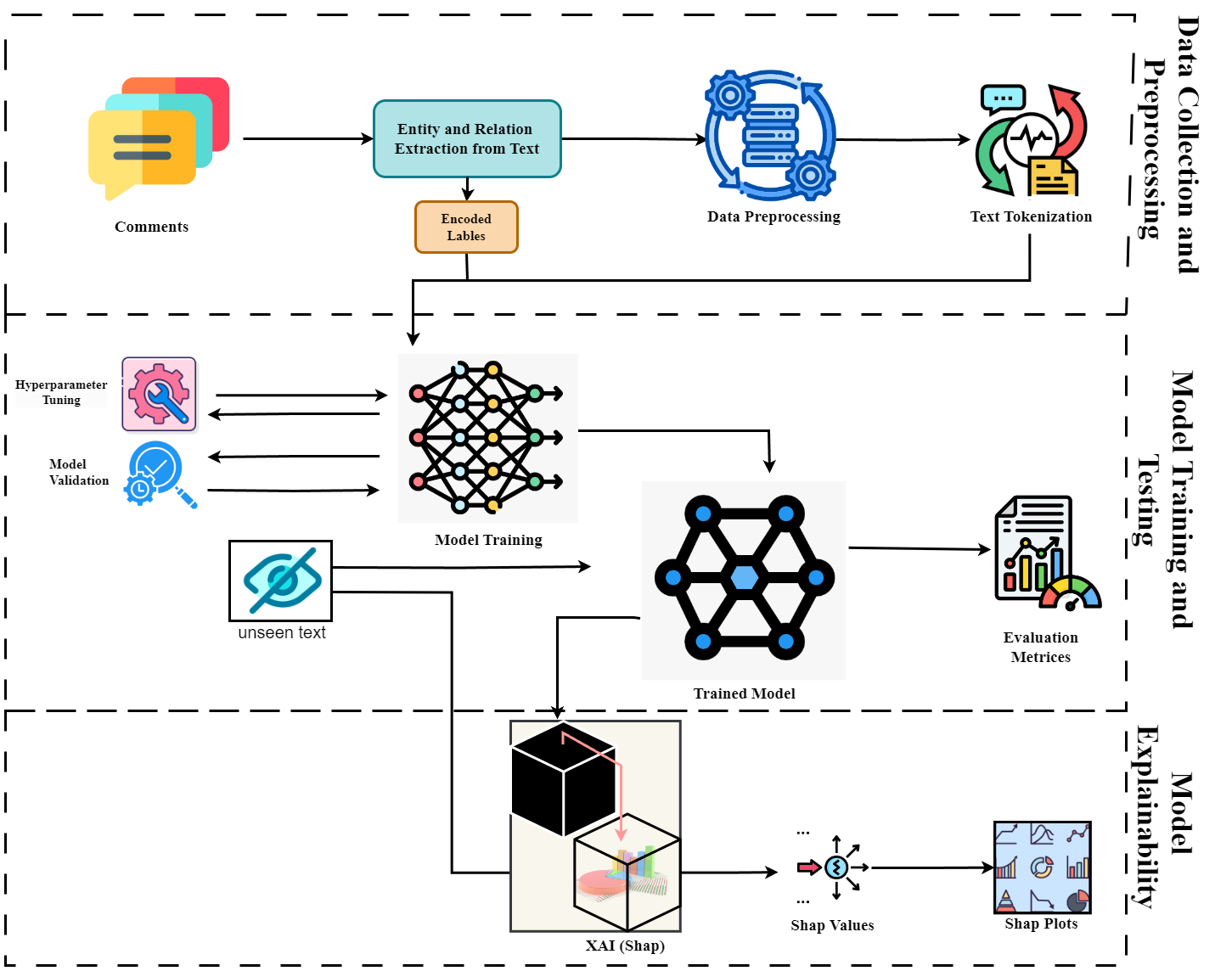}
    \caption{Workflow of the proposed approach}
    \label{woekflow}
\end{figure}
Figure \ref{woekflow} delineates the workflow diagram of the approach proposed for sentiment analysis of comments related to the HMPV virus. The entire workflow can be divided into three distinct parts. 

In the first division, a thorough and step-by-step data collection and preprocessing process is outlined. This includes extracting relevant features from the comments, preprocessing and tokenizing the text, and encoding sentiment labels (positive, negative, or neutral). These comments, typically from social media or online platforms, are cleaned and prepared to ensure the model receives high-quality data for training. 

Next, the second division focuses on model building and training. This includes essential steps like hyperparameter tuning, model selection, and evaluation using metrics like accuracy, micro and macro F1 scores. After these steps, a trained model is obtained that can predict the sentiment of previously unseen comments related to the HMPV virus.

Finally, the Explainability division applies the SHAP algorithm to the trained model using unseen data. SHAP values are computed to provide insights into the decision-making process of the model. By using the SHAP algorithm and the model’s tokenizer, the SHAP values are determined for each unseen comment, allowing us to understand how specific words or phrases in the comments influence the model’s sentiment predictions.

\section{Results}

\begin{table}[h]
    \centering
    \begin{tabular}{|l|c|c|c|c|}
        \hline
        \textbf{Model} & \textbf{Accuracy (\%)} & \textbf{Precision (\%)} & \textbf{Recall (\%)} & \textbf{F1 Score (\%)} \\
        \hline
        RoBERTa    & 89.96  & 90.12  & 89.96  & 89.97  \\
        ELECTRA    & 91.14  & 91.37  & 91.14  & 91.15  \\
        ALBERT     & 89.60  & 89.67  & 89.60  & 89.57  \\
        DistilBERT & 89.50  & 89.46  & 89.50  & 89.45  \\
        XLNet      & 93.50  & 92.50  & 93.50  & 93.00  \\
        BERT       & 91.00  & 91.00  & 91.00  & 91.00  \\
        \hline
    \end{tabular}
    \caption{Performance comparison of different transformer models.}
    \label{tab:model_performance}
\end{table}

\begin{figure}[h]
    \centering
    \includegraphics[width=1\linewidth]{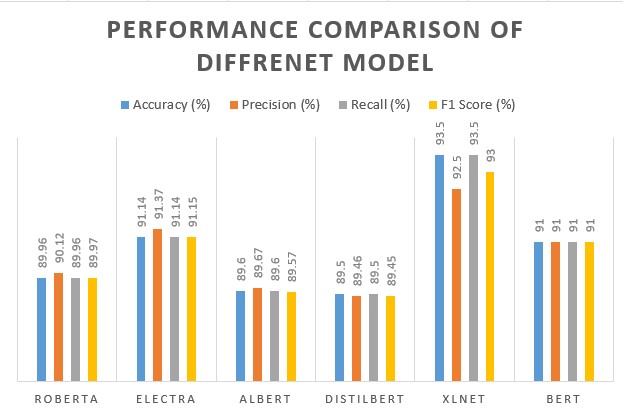}
    \caption{Comparison of Accuracy, Precision, Recall, and F1 Score for different transformer models: BERT, ELECTRA, RoBERTa, ALBERT, DistilBERT, and XLNet.}
    \label{fig:result_comparison}
\end{figure}

A comprehensive evaluation was conducted by applying various transformer-based architectures, including ELECTRA, RoBERTa, ALBERT, DistilBERT, XLNet, and BERT, to the preprocessed dataset. Each model was trained using its corresponding tokenizer, and performance was assessed using four key evaluation metrics: Accuracy, Precision, Recall, and F1 Score. The summarized results, presented in Table \ref{tab:model_performance}, highlight that XLNet achieved the best performance across the metrics.

Among the evaluated models, XLNet outperformed the others with an accuracy of \textbf{93.50\%}, followed by ELECTRA (\textbf{91.14\%}) and BERT (\textbf{91.00\%}). RoBERTa, ALBERT, and DistilBERT demonstrated lower accuracy scores, ranging from \textbf{89.50\%} to \textbf{89.96\%}. Figure \ref{fig:result_comparison} visually illustrates the comparative performance of these models, with XLNet showing significant advantages in all evaluation metrics.

\subsection{Confusion Matrix Analysis}
\begin{figure}[h]
    \centering
    \includegraphics[width=1\textwidth]{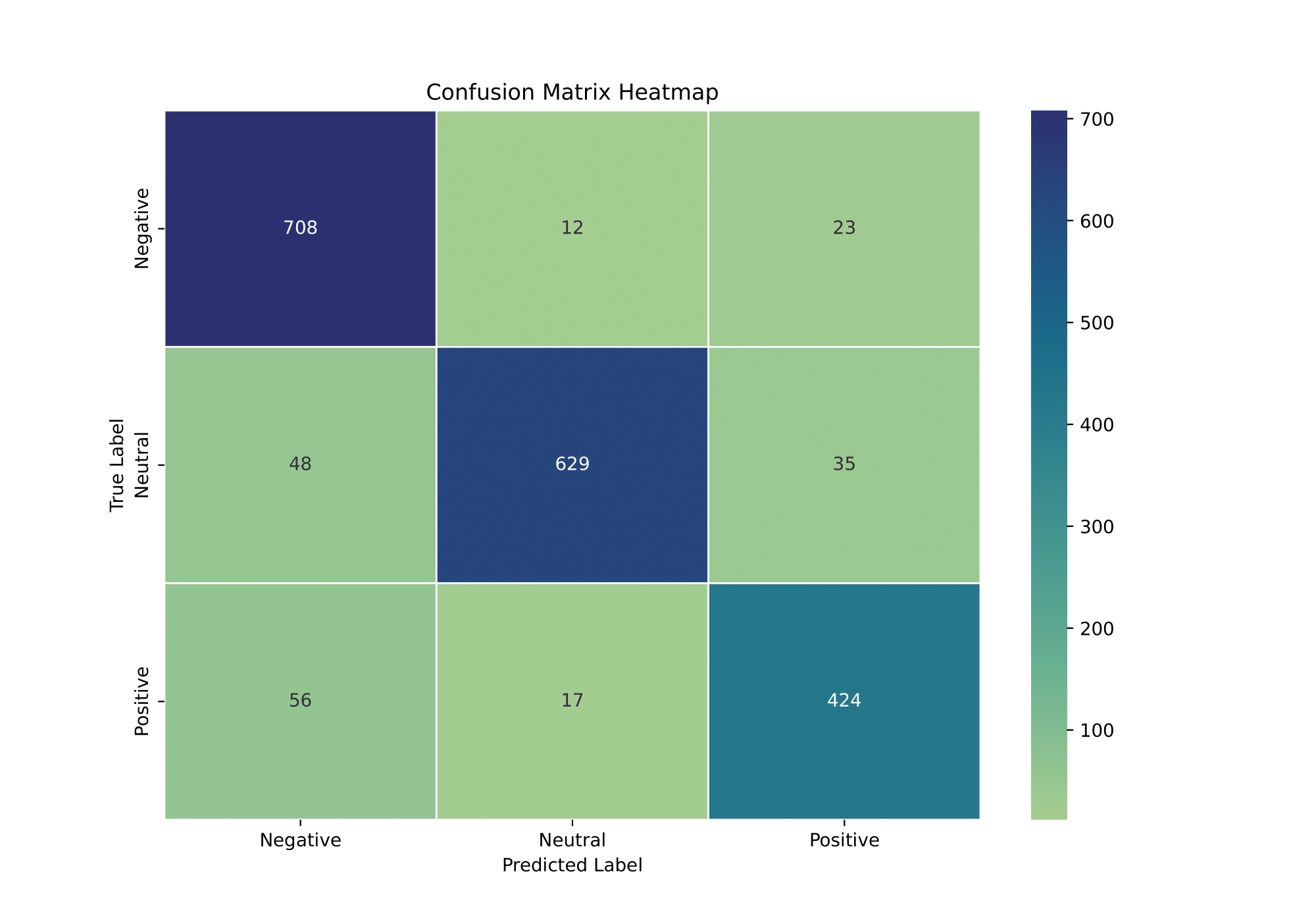}
    \caption{Confusion Matrix of the classification model. The diagonal values represent correct predictions, while the off-diagonal values indicate misclassifications.}
    \label{fig:confusion_matrix}
\end{figure}

To further analyze model performance, a confusion matrix was generated, as shown in Figure \ref{fig:confusion_matrix}. The diagonal elements of the matrix represent correct classifications, while off-diagonal elements indicate misclassifications. The model correctly classified \textbf{708 instances} of the Negative class, \textbf{629 instances} of the Neutral class, and \textbf{424 instances} of the Positive class. However, some misclassifications were observed:
\begin{itemize}
    \item \textbf{Negative Class}: 12 instances misclassified as Neutral, 23 instances misclassified as Positive.
    \item \textbf{Neutral Class}: 48 instances misclassified as Negative, 35 instances misclassified as Positive.
    \item \textbf{Positive Class}: 56 instances misclassified as Negative, 17 instances misclassified as Neutral.
\end{itemize}

These results indicate that the model performs well in distinguishing the Negative and Neutral classes, while the Positive class exhibits a moderate misclassification rate. This suggests that additional fine-tuning, data augmentation, or feature engineering may be required to improve the model’s ability to differentiate the Positive class effectively.

\subsection{Detailed Classification Report}
The confusion matrix was further analyzed to compute additional evaluation metrics, including Precision, Recall, Specificity, and Error Rate. Table \ref{tab:classification_report} presents the detailed classification report for each class, including Micro and Macro Averages.

\begin{table}[h]
    \centering
    \begin{tabular}{|l|c|c|c|c|}
        \hline
        \textbf{Class} & \textbf{Precision (\%)} & \textbf{Recall (\%)} & \textbf{F1 Score (\%)} & \textbf{Support} \\
        \hline
        Negative & 91.00 & 96.00 & 93.00 & 743 \\
        Neutral & 99.00 & 92.00 & 95.00 & 712 \\
        Positive & 88.00 & 90.00 & 89.00 & 497 \\
        \hline
        \textbf{Accuracy} & \multicolumn{3}{c}{93.5} & 1957 \\
        \textbf{Macro Avg} & 93.0 & 93.0 & 93.0 & 1952 \\
        \textbf{Weighted Avg} & 93.0 & 93.0 & 93.0 & 1952 \\
        \hline
    \end{tabular}
    \caption{Classification report showing precision, recall, and F1 score for each class.}
    \label{tab:classification_report}
\end{table}
\subsection{ROC Curve Analysis}
\begin{figure}[h]
    \centering
    \includegraphics[width=1\textwidth]{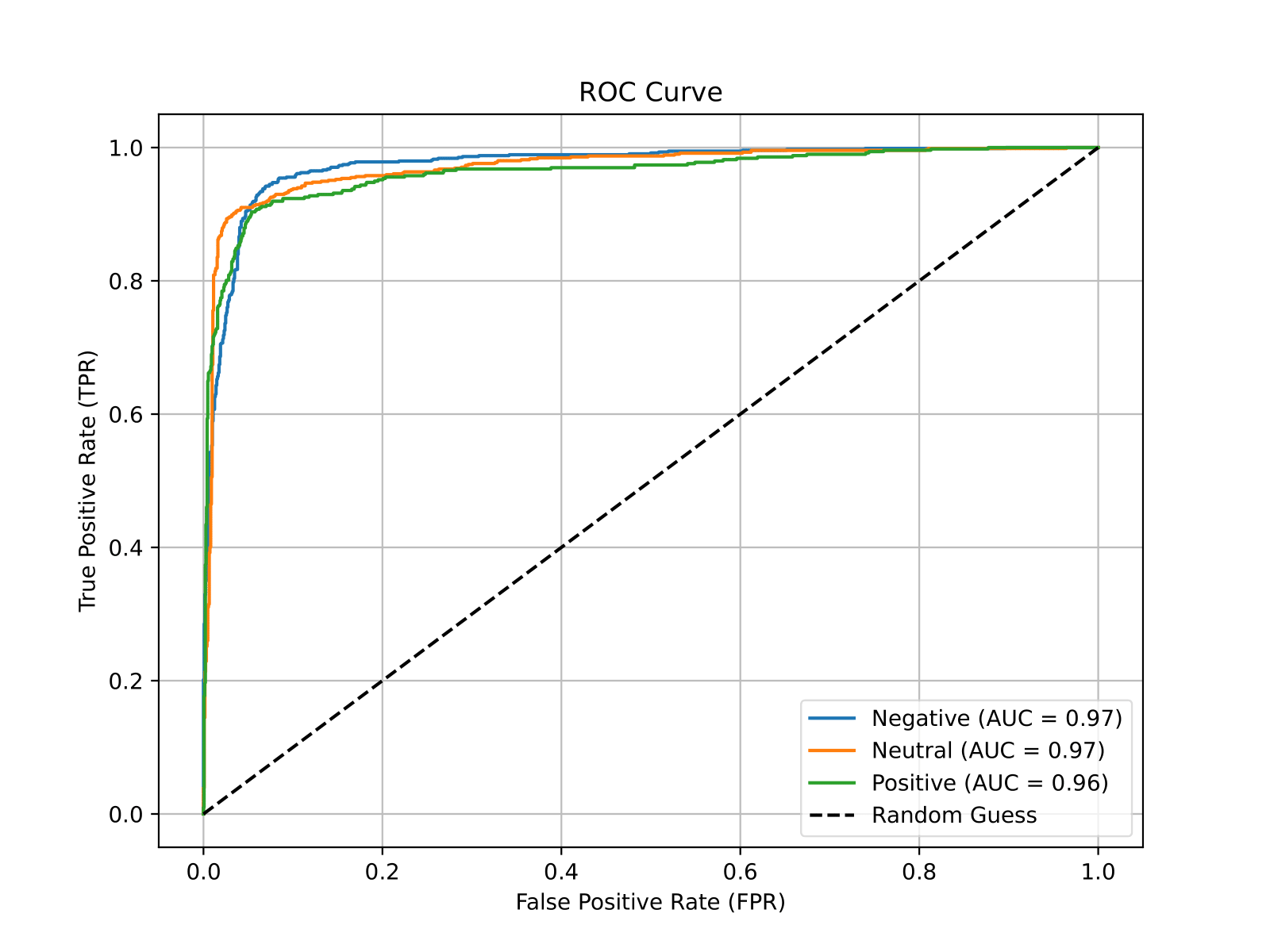}
    \caption{ROC Curve showing the performance of the classification model for different classes.}
    \label{fig:roc_curve}
\end{figure}

The Receiver Operating Characteristic (ROC) curve was plotted to assess the classification model's discriminatory power. Figure \ref{fig:roc_curve} illustrates the ROC curves for each class, along with their respective AUC (Area Under the Curve) values.

The AUC values for  Negative and Class Neutarl were both \textbf{0.97}, while Positive had a slightly lower AUC of \textbf{0.96}. These high AUC values confirm the model’s strong ability to distinguish between different sentiment categories.

\subsection{SHAP Analysis for Model Interpretability}
\begin{figure*}[h]
    \centering
    \begin{minipage}{0.5\linewidth}
        \centering
        \includegraphics[width=\linewidth]{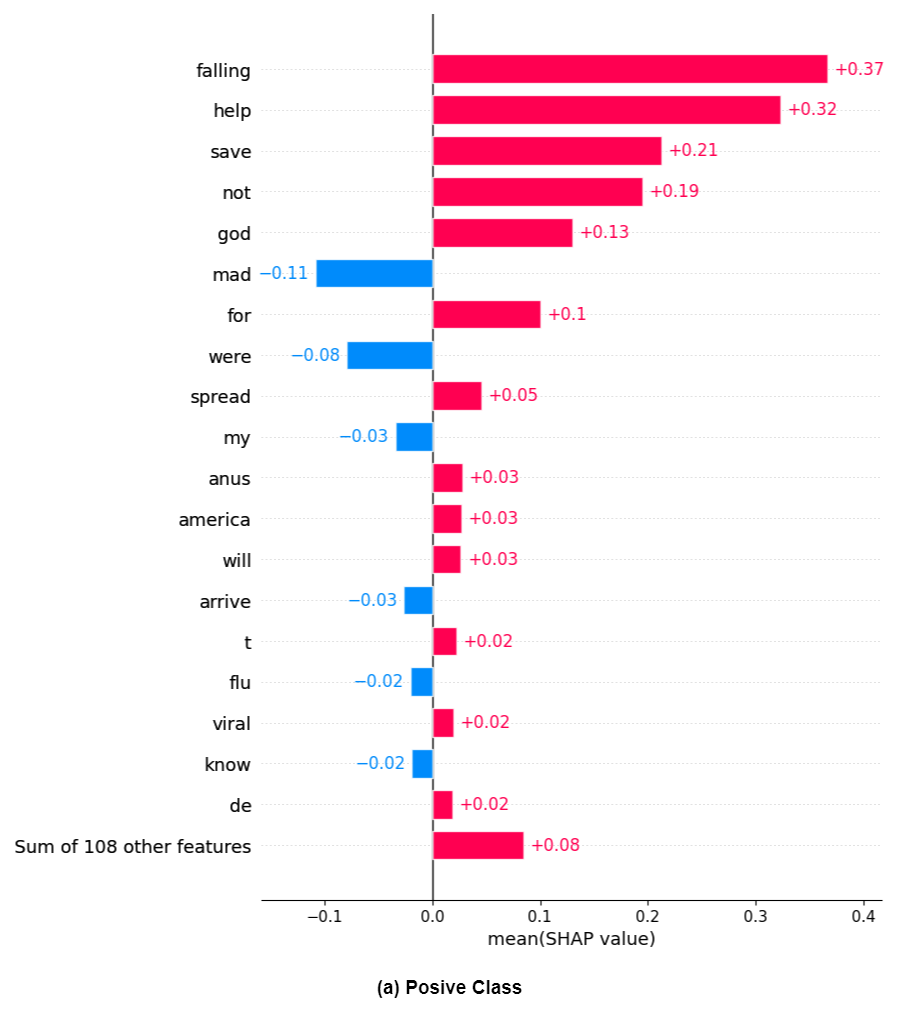}
    \end{minipage}%
    \begin{minipage}{0.5\textwidth}
        \centering
        \includegraphics[width=\linewidth]{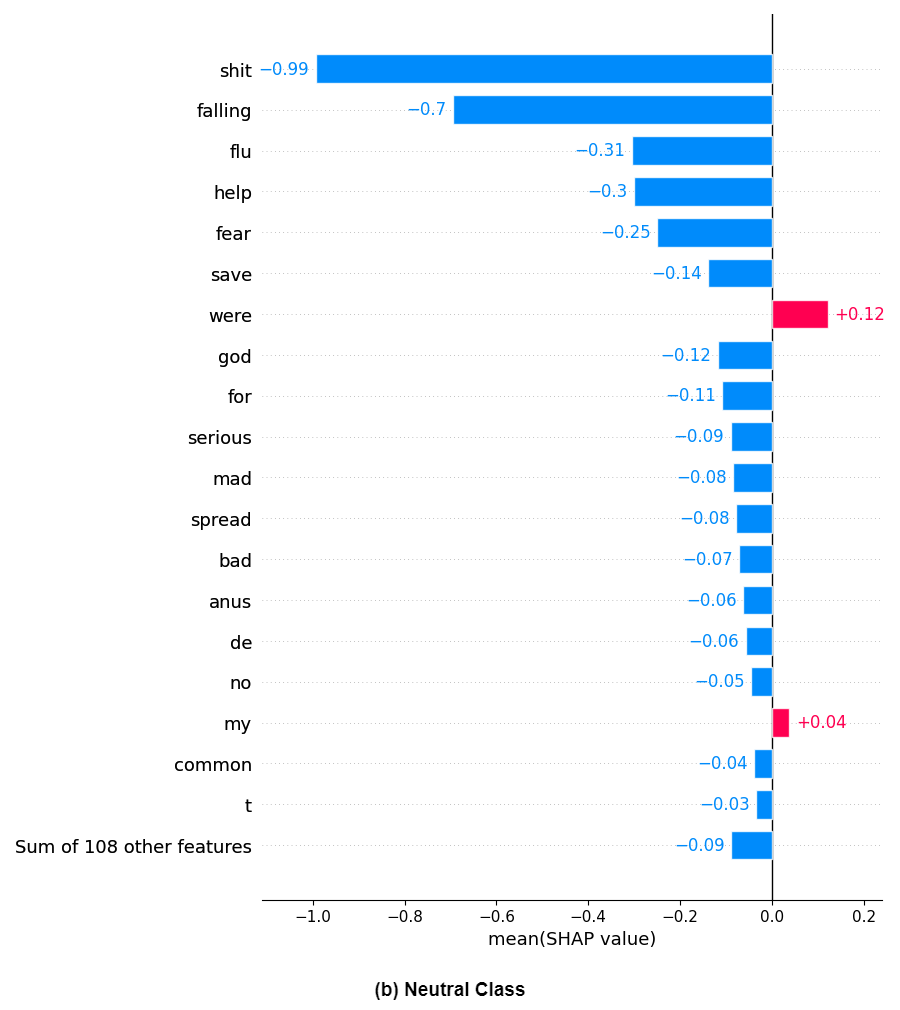}
    \end{minipage}
    
    \centering
    \begin{minipage}{0.5\textwidth}
        \centering
        \includegraphics[width=\linewidth]{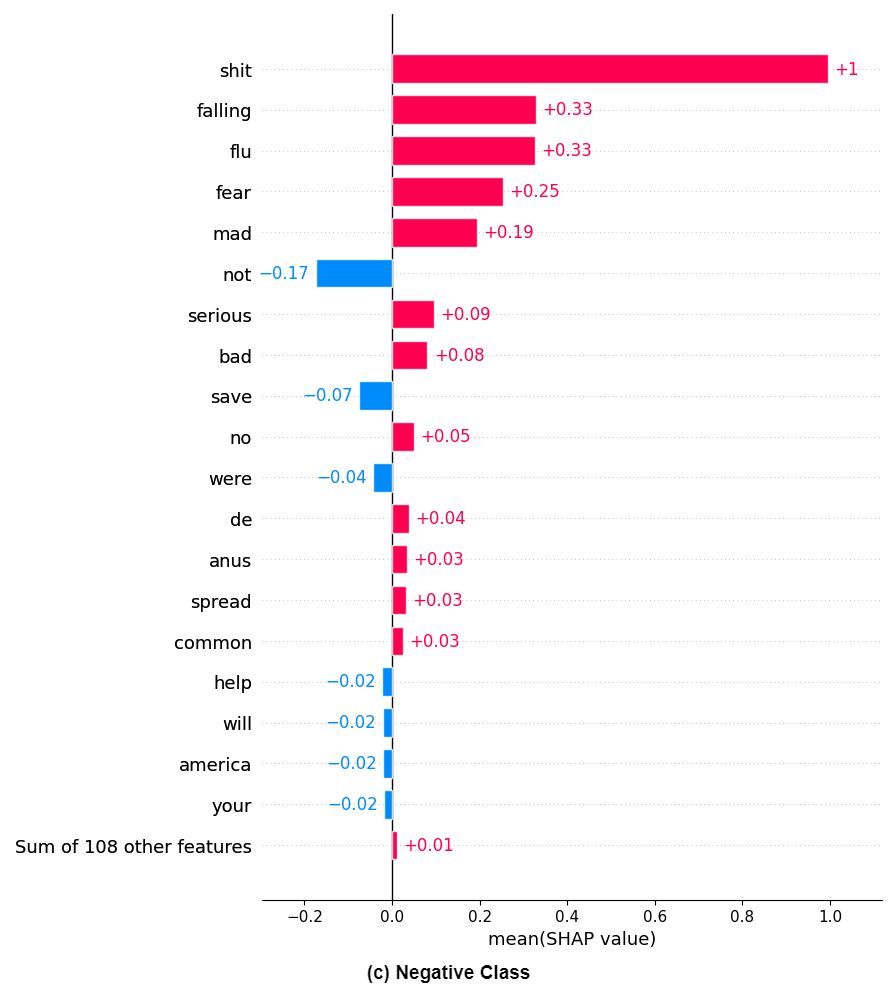}
    \end{minipage}
    
    \caption{SHAP summary plot illustrating the key features influencing model predictions.}
    \label{fig:shap_summary}
\end{figure*}
\begin{figure}[h]
        \centering
        \includegraphics[width=.8\linewidth]{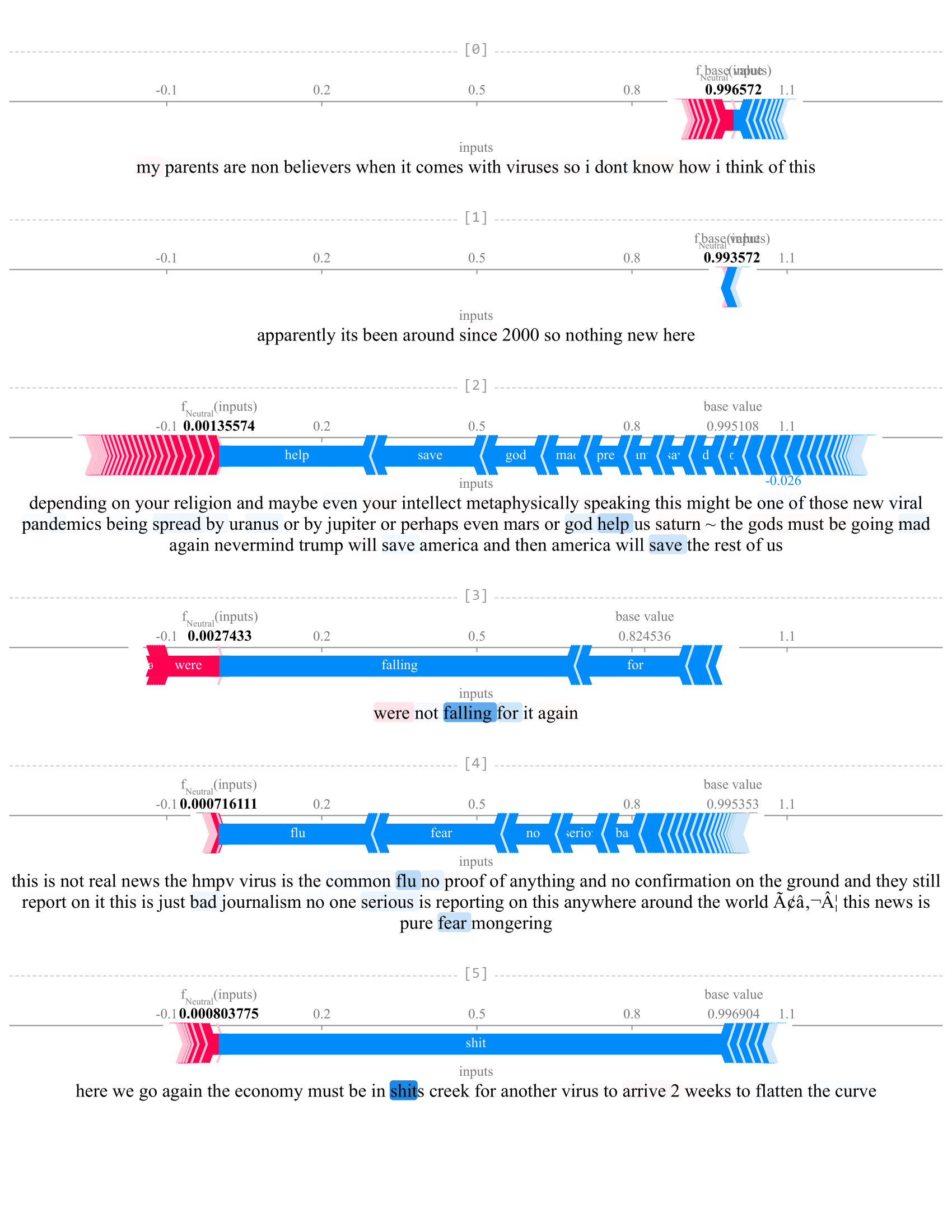}
        \caption{SHAP force plots illustrating word contributions in sentiment classification. Blue words push towards neutrality or positivity, while red words push towards negativity.}
        \label{fig:shap_analysis}
    \end{figure}
\begin{figure}[ht]
        \centering
        \includegraphics[width=.7\linewidth]{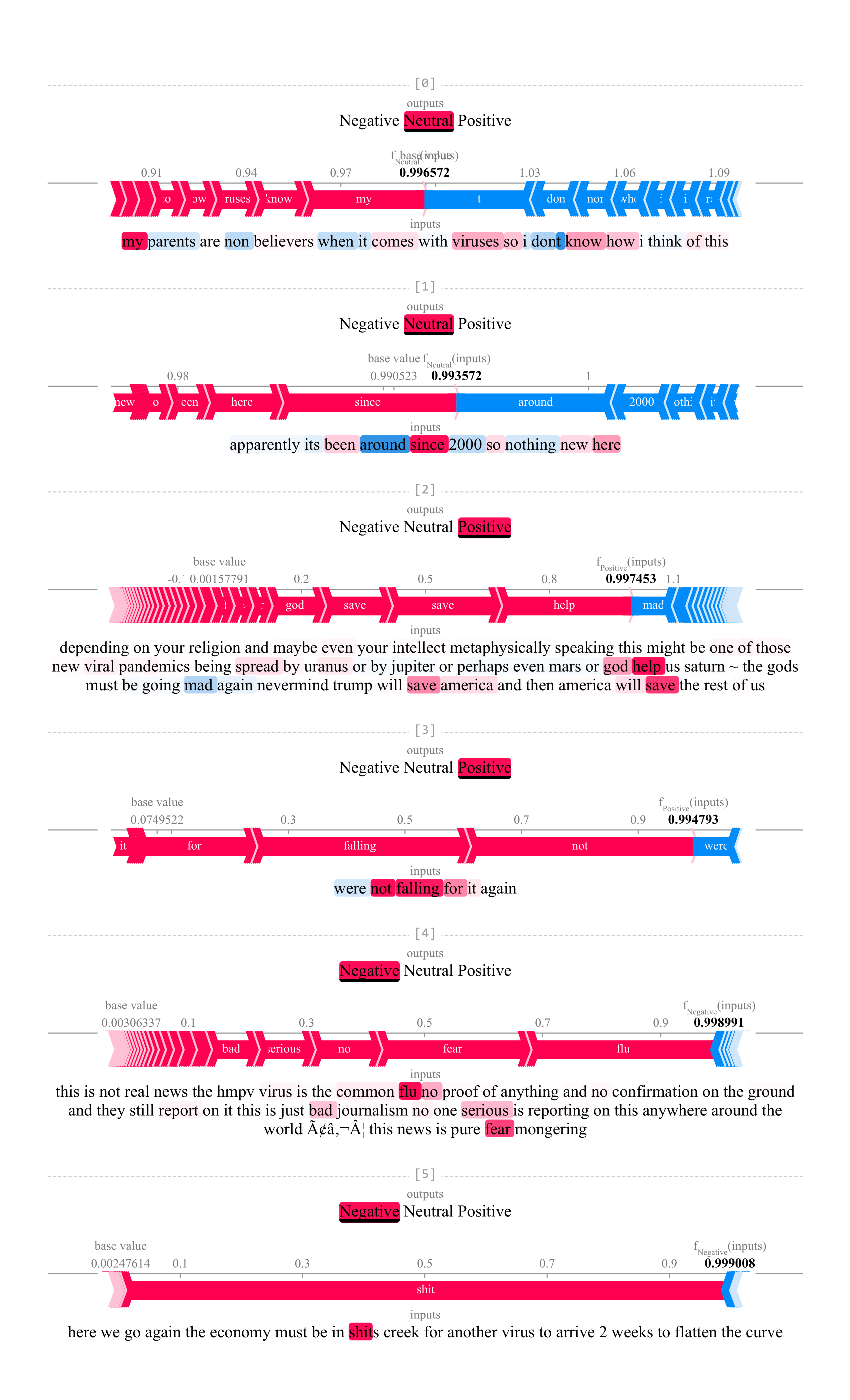}
        \caption{SHAP analysis showing the impact of individual words on model predictions. Words contributing to negative sentiment are highlighted in red, while those contributing to neutrality or positivity are marked in blue.}
        \label{fig:shap_final}
\end{figure}

To further validate XLNet as the optimal model, the SHAP (SHapley Additive exPlanations) explainer was utilized. SHAP values provide interpretability by identifying important features influencing predictions. Figures \ref{fig:shap_analysis} and \ref{fig:shap_final} present SHAP force plots, which illustrate how individual words contribute to sentiment classification. 

The SHAP analysis reveals that specific words significantly impact the model's predictions. The force plots in the figures show how each word pushes the classification toward either a positive, neutral, or negative sentiment. For example:
\begin{itemize}
    \item Words such as \textit{"flu"}, \textit{"fear"}, and \textit{"shit"} have high positive SHAP values in negative sentiment predictions, highlighting their strong contribution to negative classification.
    \item Words like \textit{"save"}, \textit{"help"}, and \textit{"mad"} exhibit high SHAP values in neutral sentiment predictions, demonstrating their influence in maintaining neutrality.
    \item Positive sentiment classifications are driven by words such as \textit{"falling"}, \textit{"save"}, and \textit{"america"}, which push the sentiment away from negativity.
\end{itemize}

The force plots also demonstrate the cumulative effect of multiple words in determining the final classification. Words with strong SHAP values in either direction directly influence whether the sentiment is classified as negative, neutral, or positive.

The experimental results demonstrate that XLNet achieves state-of-the-art performance in sentiment classification, outperforming other transformer-based models. The SHAP analysis provides interpretability by identifying key linguistic features influencing predictions. The confusion matrix, classification report, and ROC analysis further validate the model’s robustness and accuracy, confirming its effectiveness for sentiment analysis tasks.

\section{Discussion}

In this study, we developed a robust sentiment analysis model for classifying YouTube comments related to the HMPV virus, focusing on the preprocessing, model building, and explainability phases. The research highlights the challenges of handling noisy, multilingual data and demonstrates the effectiveness of advanced techniques, such as XLNet and SHAP, in overcoming these challenges.

The data preprocessing steps played a crucial role in preparing the dataset for machine learning. With a substantial portion of the initial comments containing irrelevant content, such as spam, advertisements, and errors, the preprocessing pipeline was vital in improving the quality and reliability of the dataset. Language translation, emoji translation, removal of links and special characters, stopword removal, and lemmatization were all essential steps to ensure the comments were in a consistent and analyzable format. These steps helped reduce the noise in the data, which is critical for achieving high-quality model predictions. The manual cross-checks conducted further contributed to the integrity of the data, as they ensured that only the most relevant content was retained, as shown in \ref{fig:dataPreprocess}.

The XLNet model, chosen for its ability to handle long sequences and capture bidirectional context, outperformed traditional models like BERT in sentiment analysis tasks. By leveraging permutation-based training, XLNet is capable of modeling the relationships between tokens more effectively, making it a strong candidate for analyzing informal language found in user-generated content like YouTube comments. The pretraining and fine-tuning processes allowed the model to adapt to the specific nuances of sentiment detection in online comments, and its performance on the task demonstrated the advantages of using such advanced models in the NLP space, as shown in \ref{fig:result_comparison}.

Hyperparameter tuning also played a significant role in optimizing the XLNet model. By adjusting key parameters such as learning rate, batch size, and number of epochs, we were able to improve the model's convergence speed and overall performance. The optimal hyperparameters, identified through experimentation, allowed the model to strike a balance between overfitting and underfitting, ensuring reliable predictions on unseen data. \ref{tab:hyperparameters} presents the final set of tuned hyperparameters.

The model's evaluation metrics—accuracy, micro and macro F1 scores, precision, recall, specificity, and error rate—demonstrated the effectiveness of the model in classifying sentiment accurately. The use of multiple evaluation metrics provided a comprehensive view of the model's performance, ensuring it was not biased toward a particular class. These results emphasize the importance of considering various metrics when assessing a model, especially in real-world applications where data can be imbalanced, as shown in \ref{tab:model_performance}.

One of the key innovations in this research was the application of SHAP for model explainability. By providing insights into which features of the comments influenced the sentiment predictions, SHAP enhanced the transparency of the XLNet model. This interpretability is crucial in gaining trust from users, as it helps us understand the model's decision-making process and identify potential areas for improvement. Additionally, SHAP facilitated error analysis, pinpointing the specific features that led to misclassifications. This feedback loop enables continuous model refinement, which is essential for real-world deployment, as shown in \ref{fig:shap_summary}.

Despite the promising results, there are several limitations to the current study. While XLNet demonstrated strong performance, it requires significant computational resources, especially during pretraining, which can be a barrier for smaller-scale applications. Additionally, the reliance on sentiment labeling may not always capture the full complexity of user emotions, as some comments may contain mixed sentiments or sarcasm, which are challenging to identify. Future research could explore integrating multimodal data, such as video or image content, along with text, to enhance sentiment detection in multimedia platforms like YouTube.

\section{Conclusion and Future Work}

In this study, we developed a sentiment analysis model to classify YouTube comments related to the HMPV virus, addressing the challenges of noisy, multilingual, and unstructured data. By leveraging advanced NLP techniques such as XLNet and SHAP, we demonstrated the power of state-of-the-art models in extracting meaningful insights from user-generated content. The preprocessing pipeline, which involved steps like language translation, emoji translation, and stopword removal, played a crucial role in improving the quality of the dataset and enhancing the model's performance.

Our proposed approach using XLNet for sentiment classification outperformed traditional models and achieved promising results, with high accuracy and other relevant evaluation metrics. Furthermore, the integration of SHAP allowed for enhanced model explainability, providing valuable insights into feature importance and aiding in error analysis. This combination of high-performance models and explainability techniques offers significant advantages for real-world applications, where interpretability is as crucial as accuracy.

Despite these successes, several areas require improvement. The computational cost of training large models like XLNet can be a barrier, particularly in resource-constrained environments. Additionally, the handling of mixed sentiments, sarcasm, and the incorporation of multimodal data remains a challenge that could further enhance the model's robustness and generalization. Moreover, future research could explore techniques for better handling noisy or adversarial data, ensuring that the model remains effective under varying conditions.

Looking ahead, future work will focus on optimizing the XLNet model for deployment in real-time applications. We also aim to extend the dataset to include more diverse YouTube comments, which will improve the model's generalizability and robustness. Furthermore, we plan to investigate the integration of additional features such as user metadata and video content, which could provide a more comprehensive understanding of user sentiment. Finally, exploring lightweight versions of the model for faster inference without sacrificing accuracy will be essential for real-time systems.

In conclusion, this study contributes to the growing field of sentiment analysis in social media platforms by demonstrating how advanced NLP models and explainability techniques can be used to tackle challenges in classifying user comments. Our work sets the stage for further advancements in both model performance and interpretability, paving the way for more accurate and reliable sentiment analysis systems in future applications.

\bibliographystyle{ieeetr}

\bibliography{ref.bib}



\end{document}